\pdfoutput=1

\documentclass[11pt]{article}
\usepackage{arabtex}
\usepackage{utf8}
\setcode{utf8}
\def\endabstract{\egroup}

\usepackage[final]{EACL2023}

\usepackage{times}
\usepackage{latexsym}
\usepackage[T1]{fontenc}
\usepackage[utf8]{inputenc}
\usepackage{microtype}
\usepackage{cite}
\usepackage{makecell}
\usepackage{multirow}
\usepackage{arydshln}
\usepackage{hyperref}
\usepackage{amsmath,amssymb,amsfonts}
\usepackage{algorithmic}
\usepackage{graphicx}
\usepackage{textcomp}
\usepackage{xcolor}
\usepackage{soul}
\usepackage{tablefootnote}
\usepackage{array}
\usepackage{enumitem}
\usepackage{inconsolata}
\usepackage{longtable,geometry}

\title{Arabic Synonym BERT-based Adversarial Examples for Text Classification\footnote{$^*$}}

\author{Norah Alshahrani \hspace{27pt}  Saied Alshahrani 
\hspace{27pt} Esma Wali \hspace{27pt} Jeanna Matthews \\
Department of Computer Science, Clarkson University, Potsdam, New York, USA \\
\texttt{\{\href{mailto:norah@clarkson.edu}{norah}, 
\hspace{-4pt}\href{mailto:saied@clarkson.edu}{saied}, 
\hspace{-4pt}\href{mailto:walie@clarkson.edu}{walie}, 
\hspace{-4pt}\href{mailto:jnm@clarkson.edu}{jnm}\}}\texttt{@clarkson.edu}}

\usepackage{fancyhdr}
\usepackage{lipsum}

\fancypagestyle{specialfooter}{%
  \fancyhf{}
  
  \fancyfoot[L]{*Accepted at EACL2024: Student Research Workshop.}
}

\begin{document}

\thispagestyle{specialfooter}

\maketitle

\begin{abstract}

Text classification systems have been proven vulnerable to adversarial text examples, modified versions of the original text examples that are often \textcolor{black}{unnoticed} by human eyes, yet \textcolor{black}{can} force text classification models to \textcolor{black}{alter their classification}. Often, \textcolor{black}{research} work\textcolor{black}{s} \textcolor{black}{quantifying the impact of adversarial text attacks \textcolor{black}{have} been applied only to models trained in English}. In this paper, we introduce the \textcolor{black}{\emph{first}} \textcolor{black}{word-level study of adversarial attacks in Arabic. \textcolor{black}{Specifically}, we use} a synonym (word-level) attack using a Masked Language Modeling (MLM) task with a BERT model in a black-box setting to assess the robustness of the state-of-the-art text classification models to adversarial attacks in Arabic. To evaluate the grammatical and semantic similarities of the newly produced adversarial examples using our synonym BERT-based attack, we invite four human evaluators to assess and compare the produced adversarial examples with their original examples. We also study the transferability of these newly produced Arabic adversarial examples to various models and investigate the effectiveness of defense mechanisms \textcolor{black}{against these adversarial examples on the BERT models}. We find that \textcolor{black}{fine-tuned} BERT models were more susceptible to 
\textcolor{black}{our synonym}
attacks than the \textcolor{black}{other} Deep Neural Networks (DNN) models \textcolor{black}{like} \textcolor{black}{WordCNN and WordLSTM} \textcolor{black}{we trained}. We also find that \textcolor{black}{fine-tuned} BERT models were more susceptible to \textcolor{black}{transferred }attacks. We, lastly, find that \textcolor{black}{fine-tuned} BERT models successfully \textcolor{black}{regain at least 2\% in accuracy} after applying adversarial training as an initial defense mechanism.
\end{abstract}


\section{Introduction}
\label{sec:1}

Machine Learning models, in general, are vulnerable to adversarial attacks, which are small, crafted perturbations done usually by altering the original input of these models \textcolor{black}{in order to change their classification} \textcolor{black}{\citep{57, 62, 60, 63, 61, 58, 59}}.

\begin{figure}[!ht]
    \centering
    \includegraphics[width=\linewidth]{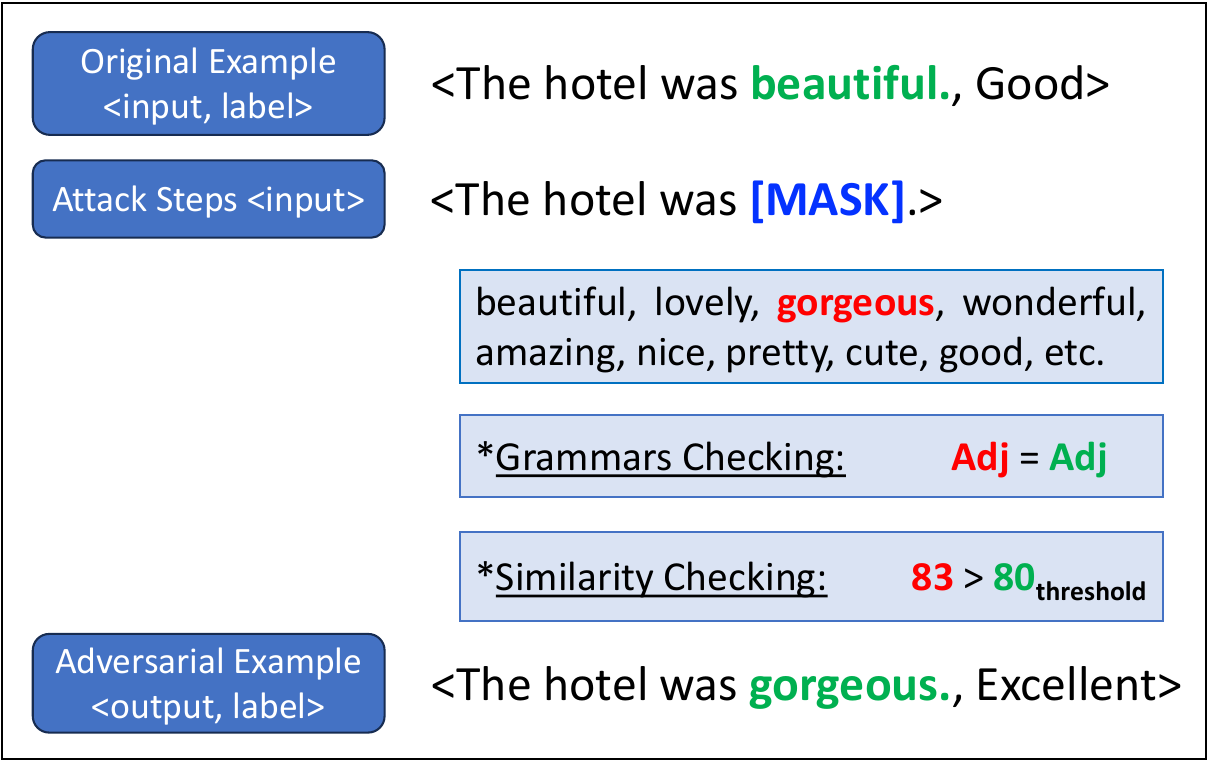}
    \caption{A diagram illustrates our attack steps for generating synonym adversarial attacks using an MLM task with BERT. The attack first predicts synonym tokens and then checks their grammar and semantic similarities. Once a predicated token satisfies the grammar and semantic checkings, we have an attack candidate example.} 
    \label{fig:1}
    \vspace{-15pt}
\end{figure}

 Research on adversarial attacks is often in the domain of image classification systems \citep{15, 17, 18, 19, 49, 50} or speech recognition systems \citep{48,45,44,22,47, 46}. For example, in the domain of image classification, the images are the original inputs, and the attackers could negatively affect the performance of these systems by introducing small perturbations to the input images \citep{15, 17, 18, 19, 27}.  
 Examining adversarial attacks in the domain of Natural Language Processing (NLP) can be \textcolor{black}{especially} challenging due to 
 the discrete nature of the input texts and the requirement to preserve both semantic coherence and grammatical correctness with the original texts \citep{32, 65, 64}. 
 
 Generally, research studies of adversarial attacks can be classified as white-box, gray-box, or black-box. In a white-box approach, attackers can fully access the model architecture, weights, parameters, or training datasets \citep{24}, whereas in the gray-box approach, the attackers have limited access to the model architecture \citep{33}. Finally, in the black-box approach, the attackers cannot access the model architecture but only query the model and get a prediction in return \citep{23,34}.

A few common techniques for producing adversarial text examples have been addressed widely in the NLP field, such as character-level attacks (like inserting, removing, or swapping one or more characters within a word), word-level attacks (like inserting, removing, or replacing a word), and sentence-level attacks (like inserting, removing, or replacing a word or more than a word in a sentence) \citep{26, 24,25,27,28}. Some of these techniques can result in unnatural adversarial examples, making them easily distinguishable by humans, but recent research indicates that using rule-based synonym replacement strategies could generate adversarial text examples that appear more natural and similar to the original examples \citep{29,30,1,2,31}.

Only \emph{two} papers addressed adversarial text attacks in Arabic, and both are character-level. \citet{3} proposed character-level adversarial attacks that rely on changing the morphological form of adjectives by adding one or more characters, which violates the noun-adjective agreement. An adjective is a word that describes a noun, and it must agree with the noun in definiteness (i.e., definite or indefinite), number (singular or dual), and gender (i.e., feminine or masculine).  \citet{4} also proposed character-level attacks, which relied on the flip of one or two Arabic characters chosen based on non-native Arabic learners' most common spelling mistakes (usually incorrect use of visually similar characters). Both of these studies considered only the character-level adversarial examples and did not investigate the impact of transferability of these attacks among targeted models nor the effectiveness of defense mechanisms like adversarial training.

In this paper, we introduce the \emph{first} 
study of word-level adversarial attacks in Arabic. We develop synonym-based word-level attacks using a Masked Language Modeling (MLM) task with an Arabic BERT model in a black-box manner against three state-of-the-art sentiment analysis classifiers/models: BERT (Bidirectional Encoder Representations from Transformers) \citep{8}, WordCNN (word-based Convolutional Neural Networks) \citep{7}, and WordLSTM (Word-based Long Short-term Memory) \citep{6}. We train these models on two available and large Arabic datasets, HARD (Hotel Arabic Reviews Dataset) \citep{5} and MSDA (Sentiment Analysis for Social Media Posts in Arabic Dialect) \citep{12}, to automatically generate adversarial text examples, attack these models using those generated adversarial text examples, and finally, assess the robustness of these models against adversarial text examples. 

We also use human evaluation to evaluate the newly produced adversarial text examples using two criteria: grammatical similarity and semantic similarity. Furthermore, we study the transferability of these adversarial text examples generated by various models on different models studied and deeply investigate the effectiveness of the adversarial training defense mechanism on the BERT models against these adversarial text examples. Figure \ref{fig:1} illustrates the process of generating synonym-based word-level adversarial examples generation which we discuss in detail in Section \ref{sec:3}.

Section \ref{sec:2} reviews related research works, while Section \ref{sec:3} delves into the methodology in more detail. Sections \ref{sec:4} and \ref{sec:5} discuss the automatic and human evaluations and their results. In Sections \ref{sec:6} and \ref{sec:7}, we describe the transferability and defense mechanisms, along with their results. Lastly, in Sections \ref{sec:8} and \ref{sec:9}, we conclude our paper by addressing its limitations and summarizing our contributions.

\section{Related Work}
\label{sec:2}
Adversarial text attacks have been widely studied in the NLP field from different perspectives, like the attack setting (white-box, gray-box, and black-box), targeted space (embedding or input), and attack method (character-level, word-level, sentence-level) \citep{51,26,24, 25, 27, 28, 29,39, 30,31, 43,40, 1, 36}. Here, we will focus on related work that shares the same methodology as ours, taking advantage of the pre-trained models like the BERT model and its MLM training objective, but notably, they are all done only in English.

\citet{1} \textcolor{black}{integrated two synonym replacement strategies  \citep{29,30} and} proposed a baseline method called \textsc{TextFooler} to efficiently generate adversarial examples using synonym replacement techniques through word embeddings, ensuring the preservation of similar semantic meaning compared to the original words. Their findings show that \textcolor{black}{pre-trained} BERT models and \textcolor{black}{other} Deep Neural Networks (DNN) models were vulnerable to these adversarial text attacks, which could lead to misclassification or incorrect textual entailment predictions. The authors also emphasized improving the robustness of NLP models by incorporating defense mechanisms and testing their performance against adversarial attacks.

\textcolor{black}{Several studies have proposed novel synonym replacement techniques using the MLM task with Large Language Models (LLMs) like BERT. For instance,}  \citet{2} proposed a novel method called BERT-based Adversarial Examples (BAE) for generating adversarial examples for text classification using MLM with BERT models. By utilizing pre-trained BERT’s ability to capture semantic meaning and context in text, BAE could generate adversarial examples more effectively than previous methods, avoiding detection by state-of-the-art classifiers. The authors also evaluated the effectiveness of BAE against various text classification models and demonstrated that it could generate robust adversarial examples. 
\citet{31} proposed a practical method called BERT-Attack using MLM with BERT models to predict sub-word expansion, ensuring the generation of fluent and semantically preserved adversarial text examples. These adversarial examples successfully fooled the state-of-the-art models, such as the fine-tuned BERT models for various downstream tasks in NLP in a black-box manner. The authors evaluated the BERT-Attack’s effectiveness against BERT models in various text classification tasks and demonstrated its ability to reduce the accuracy of these models significantly. \textcolor{black}{Lastly,} \citet{36} proposed an attack method called CLARE, a contextualized adversarial example generation model that could generate fluent and grammatically accurate outputs through a fill-in-mask procedure using MLM with RoBERTa models. CLEAR introduced three contextualized perturbations: replace, insert, and merge, which permit generating outputs of varying lengths, where it could flexibly integrate these perturbations and apply them at any position in the inputs and then use them to attack the BERT models. The authors finally evaluated CLARE's effectiveness against  BERT models and demonstrated that CLARE achieved the best performance with the least modifications by combining all these three perturbations.

\section{Methodology}
\label{sec:3}
\subsection{Datasets Used}
In this work, we select two large Arabic datasets designed for text classification tasks: hotel reviews and sentiment analysis. We purposely chose one dataset written mostly in Modern Standard Arabic (MSA) and another written in Dialectical Arabic (DA) to observe how the models would behave when trained on different Arabic dialects. 

\hspace{-11pt}$\bullet$ \textbf{Hotel Arabic Reviews Dataset (HARD)} is a balanced dataset with  93K hotel reviews written mostly in MSA, collected from Booking.com, and has four classes \citep{5}, which we remap to Poor, Fair, Good, and Excellent, instead of the original numerical labels (1, 2, 4, and 5).\footnote{Originally, labels 1 and 2 were negative, and 4 and 5 were positive. Notably, users were not given the choice of 3, only 5, 4, 2, or 1. We found this labeling confusing and remapped simply 1 to Poor, 2 to Fair, 4 to  Good, and 5 to Excellent.}

\hspace{-11pt}$\bullet$  \textbf{Sentiment Analysis for Social Media Posts in Arabic Dialect (MSDA)} is a balanced dataset that includes  50K posts written mostly in DA, collected from the $\mathbb{X}$ platform (formerly Twitter), and has three classes \citep{12}:  Positive,  Neutral, and  Negative. 


\begin{table}[!ht]
\centering
\begin{tabular}{cccc} 
 \hline
 \textbf{Dataset} & \textbf{Avg Length} & \textbf{STD} & \textbf{Max Length}\\ 
  \hline
 HARD & 19.50 &  19.77 &503 \\ 
  \hline
 MSDA & 9.99 & 9.46 & 326\\ 
 \hline
\end{tabular}
\caption{\label{tab:1}
The general statistics of the two used datasets in terms of the Average Length (\#words), Standard Deviation (STD), and Maximum Length (\#words).}
\end{table}

\subsection{ Models \textcolor{black}{Targeted}}
We train three deep learning classifiers/models \textcolor{black}{ that are widely used for text classification tasks}: WordLSTM (Word-based Long Short-term Memory) \citep{6}, WordCNN  (Word-based Convolutional Neural Network) \citep{7}, and BERT (Bidirectional Encoder Representations from Transformers) \citep{8}, on the HARD and MSDA datasets.

We use the same hyperparameters as \citet{1} used for the models in our study. \textcolor{black}{We train WordCNN and WordLSTM models from scratch.} For WordCNN models \citep{7},  we use three window sizes of 3, 4, and 5, and 100 filters for each window size. For the WordLSTM models, we use one bidirectional LSTM layer with 150 hidden states \citep{6}. We train a GloVe (Global Vectors for Word Representation) model on each dataset to generate word vectors of size 200 dimensions for both models \citep{35} and use these GloVe embedding words to train the embedding layers in the WordCNN, and WordLSTM models on each dataset. \textcolor{black}{For the BERT model, we begin with a pre-trained Arabic BERT model called AraBERT$_{\textsc{BASE}}$ v2\footnote{AraBERT$_{\textsc{BASE}}$ v2 model can be accessed here: \href{https://huggingface.co/aubmindlab/bert-base-arabertv2}{https://huggingface.co/aubmindlab/bert-base-arabertv2}.} \citep{9}, which has 12 layers with 768 hidden states, 12 heads, resulting in 136M trainable parameters, and we then fine-tune for text classification tasks using each dataset.} \textcolor{black}{We use different levels of data preprocessing before training WordLSTM and WordCNN models and fine-tuning BERT models due to the differences in their architectures, like} \textcolor{black}{the utilization of contextual embeddings in BERT models.} 

Table \ref{tab:2} shows the original accuracy (evaluation accuracy) of each model on each dataset. BERT models score the highest evaluation accuracies: 83\% and 86\% on HARD and MSDA, respectively. The WordLSTM model is not far behind, but the WordCNN models perform substantially worse. 

\begin{table}[!ht]
\centering
\begin{tabular}{cccc} 
 \hline
 \textbf{Dataset} & \textbf{WordCNN} & \textbf{WordLSTM} & \textbf{BERT}\\ 
  \hline
 HARD & 75\% & 80\% & 83\% \\ 
  \hline
 MSDA & 77\% & 83\% &  86\%\\ 
 \hline
\end{tabular}
\caption{\label{tab:2}Original accuracy (evaluation accuracy) of each model on each dataset \textcolor{black}{(a test set of 10\% of each dataset)}.}
\end{table}

\vspace{-10pt}

\subsection{Adversarial Text Generation}
The adversarial text generation task involves working with a dataset $D$ in the form of ($X$, $Y$), composed of pairs examples $X$ and labels $Y$ in the form of \{($x_1$, $y_1$), ..., ($x_n$, $y_n$)\}, alongside a black-box classifier $C$: $X$ $\rightarrow$ $Y$. We assume a soft label (with probability score) in a black-box setting, where the attacker can only query the classifier $C$ for output labels $Y$ and probabilities $P$ given specific inputs, without access to any of the model’s parameters, weights, gradients, architecture, or training data.

Given an input example $x$, composed of $W$ words in the form of ($x$ = [$w_1$,$w_2$,$w_3$, ..., $w_n$], $y$), our goal is to create adversarial examples $X_{\textsc{ADV}}$, in such a way that $C$($X_{\textsc{ADV}}$) $\neq$ $Y$, meaning the prediction labels $Y$ of $C$($X$) do not equal the adversarial labels $Y_{\textsc{ADV}}$ returned by $C$($X_{\textsc{ADV}}$). Moreover, we aim for the adversarial examples $X_{\textsc{ADV}}$ to exhibit grammatical correctness and maintain semantic similarity to the original inputs $X$.

\subsubsection{Adversarial Text Generation Steps}
Here, we present the steps we use for generating adversarial examples $X_{\textsc{ADV}}$. We randomly select 1000 samples (original examples $X$) from each dataset and process them following these steps:\\

\vspace{-10pt}

\hspace{-12pt}\textbf{1) Word Importance Ranking:}
We use the same scoring function  ($I_{w_i}$) as \citet{1} to measure the influence of a word $w_i$. Specifically, we quantify the importance of each token/word $w_i$ in a sentence by deleting the tokens  (one token each at a time)  and calculating the prediction scores’ change before and after deleting that word $w_i$. 
We also 
clean the input original example $x$ by removing the noise, emojis, stopwords, and punctuation marks using the \texttt{NLTK} Python library\footnote{Natural Language Toolkit (NLTK): \href{https://www.nltk.org/}{https://www.nltk.org}.} before we feed the example $x$ to the scoring function  ($I_{w_i}$) to reduce the computational overhead and ensure only words are fed to the scoring function ($I_{w_i}$). \\

\vspace{-10pt}

\hspace{-12pt}\textbf{2) Word Replacement Strategy:}
We repetitively replace the most important words (one important word $I_{w_i}$ at a time) in the input original example $x$ using the MLM task with the BERT model to find synonym words for that important word $I_{w_i}$. 

We use \textcolor{black}{a different pre-trained version of}  AraBERT$_{\textsc{BASE}}$ v02\footnote{AraBERT$_{\textsc{BASE}}$ v02 model can be accessed here: \href{https://huggingface.co/aubmindlab/bert-base-arabertv02}{https://huggingface.co/aubmindlab/bert-base-arabertv02}.} \textcolor{black}{as our MLM model} \citep{9} to generate the synonym words $W_s$ and assigned the top $K$ value to 50. After that, we use the CAMeLBERT-CA POS-EGY model\footnote{CAMeLBERT-CA POS-EGY model can be accessed here: \href{https://huggingface.co/CAMeL-Lab/bert-base-arabic-camelbert-ca-pos-egy}{https://huggingface.co/CAMeL-Lab/bert-base-arabic-camelbert-ca-pos-egy}.} \citep{10} as our Part-of-Speech  (POS) tagger to ensure that the generated synonym words are grammatically correct. Next, we use a Sentence-Transformers model\footnote{Sentence-Transformers model can be accessed here: \href{https://huggingface.co/sentence-transformers/paraphrase-multilingual-mpnet-base-v2}{https://huggingface.co/sentence-transformers/paraphrase-multilingual-mpnet-base-v2}.} to ensure the newly generated examples with the replaced synonym words are semantically similar to the original examples. We specifically used a multilingual pre-trained MPNet (Masked and Permuted Pre-training for Language Understanding) model \textcolor{black}{\citep{52}} trained on parallel data for 50+ languages, including Arabic \citep{11}. We calculate the similarity score using the cosine similarity metric and set the similarity threshold to 0.80, as set by \citet{1}. After the newly generated example passes the POS and similarity checkers, we finally have an attack candidate example derived from the original example $x$.\\

\vspace{-10pt}

\hspace{-12pt}\textbf{3) Synonym \textcolor{black}{BERT-based} Attack:}
Finally, we call an attack candidate example an `adversarial example' ($x_{adv}$) if 
it flips the prediction label $y$ (the prediction label of the original example $y$ before the attack $\neq$ the prediction label of the attack candidate example $y_{adv}$ after the attack). In other words, after replacing a word $w$ with its synonym word $w_s$ in the original example $x$ \textcolor{black}{using our synonym attack}, \textcolor{black}{the} adversarial example $x_{adv}$ force the targeted \textcolor{black}{model/}classifier to change \textcolor{black}{its} classification.\footnote{Appendix \hyperlink{page.9}{A} provides some concrete examples of Arabic adversarial text examples generated using  our synonym attack, along with their labels and their English translations.}

\vspace{-4pt}
\section{Automatic Evaluation}
\label{sec:4}

\subsection{ Evaluation Metrics}
\label{sec:4.1}
We evaluate our synonym BERT-based attack using four metrics: Attack Success Rate (Att.$_{\textsc{SR}}$), Accuracy Before Attack (Acc.$_{\textsc{BA}}$), Accuracy After Attack (Acc.$_{\textsc{AA}}$),  and Attack Decrease Rate (Att.$_{\textsc{DR}}$). 

\vspace{5pt}

\hspace{-12pt}\textbf{1) Attack Success Rate (Att.$_{\textsc{SR}}$)} is a metric designed to measure the successfulness of \textcolor{black}{our synonym attack} on a specific model and dataset, and it is calculated by dividing the number of adversarial examples produced by a model from a dataset by the total number of the randomly selected samples (original examples) multiplied by 100 for percentage normalization.\\

\vspace{-10pt}
\hspace{-12pt}\textbf{2) Accuracy Before Attack  (Acc.$_{\textsc{BA}}$)} is calculated by taking the mean of prediction scores of the total number of selected samples input to the targeted model in a black box setting, meaning we only use the prediction scores instead of the targeted model's original accuracy (evaluation accuracy). We believe taking the original accuracy of the model here is not a black-box manner because attackers are not supposed to know anything about the targeted model, including its original accuracy. \\

\vspace{-10pt}

\hspace{-12pt}\textbf{3) Accuracy After Attack  (Acc.$_{\textsc{AA}}$)} is calculated by taking the mean of prediction scores of the total number of selected samples input to the targeted model in a black box setting after applying \textcolor{black}{our synonym attack.}\\

\vspace{-10pt}

\hspace{-12pt}\textbf{4) Attack Decrease Rate (Att.$_{\textsc{DR}}$)} is a metric designed to measure the effectiveness of \textcolor{black}{our synonym attack} on a specific model and dataset, and it is calculated simply by taking the difference between the Accuracy Before Attack (Acc.$_{\textsc{BA}}$) and Accuracy After Attack (Acc.$_{\textsc{AA}}$).

\subsection{Evaluation Results}
\textcolor{black}{We first choose 1000 randomly selected examples 
form each dataset: HARD and MSDA, following \citet{1} and \citet{36}. Next,} we evaluate our proposed synonym 
attack using metrics defined in subsection \ref{sec:4.1} above. The results of the automatic evaluations of our attack on each targeted model (WordCNN, WordLSTM, and BERT) and each dataset (HARD and MSDA) are displayed in Table \ref{tab:3}. We find that our attack has successfully decreased the accuracies of the targeted models on each dataset, measured by the Acc.$_{\textsc{BA}}$ and Acc.$_{\textsc{AA}}$ metrics. For the 1000 randomly chosen examples, the BERT models \textcolor{black}{fine-tuned} on the MSDA and HARD datasets scored 90.55\% and 88.59\% as accuracies before our attack,  and their accuracies after our attack have dropped to 63.62\% and 73.90\% on both datasets, respectively. 
\textcolor{black}{It is clear that our} attack strategy of substituting a word with 
a synonym works well because it creates a new example (adversarial example) that the targeted model has not encountered or seen before, forcing the targeted model to misclassify and cause a drop in its accuracy after the attack. \textcolor{black}{Yet,} \textcolor{black}{we believe that the number of the evaluated examples, the different levels of data preprocessing, and the prediction misclassification rate of the block-box models} \textcolor{black}{are \textcolor{black}{possible interpretations} of the \textcolor{black}{noticeable} difference between the original accuracy mentioned in Table \ref{tab:2}  and the Acc.$_{\textsc{BA}}$ in Table \ref{tab:3}. Regardless \textcolor{black}{of these constraints}, our results demonstrate that all three models are indeed susceptible to our synonym attacks.}



\begin{table*}[ht]
\centering
\tiny\renewcommand{\arraystretch}{0.90}
\resizebox{\textwidth}{!} {
\begin{tabular}{ccccccc} 
 \hline
 \multirow{2}{*}{\textbf{Metric}} & \multicolumn{2}{c}{ \textbf{WordCNN}}  & \multicolumn{2}{c}{\textbf{WordLSTM}} & \multicolumn{2}{c}{\textbf{BERT}}\\ 
  \cline{2-7}
   & \textbf{HARD} & \textbf{MSDA} & \textbf{HARD} &  \textbf{MSDA} &  \textbf{HARD} &  \textbf{MSDA}\\ 
\hline
Attack Success Rate (Att.$_{\textsc{SR}}$)& 50.00\% & 30.00\% &  51.00\% & 25.00\% & 51.00\% & 26.00\%\\ 
  \hline
Accuracy Before Attack (Acc.$_{\textsc{BA}}$) & 32.09\% & 45.15\% & 34.82\% & 47.48 \% & 88.59\% & 90.55\%   \\ 
  \hline
Accuracy After Attack (Acc.$_{\textsc{AA}}$) & 32.05\% & 39.31\% &  33.90\% & 41.73\% & 73.90\% & 63.62\% \\ 
 \hline
Attack Decrease Rate (Att.$_{\textsc{DR}}$) & 00.04\% & 05.84\% &  00.92\% & 05.75\% &  14.69\% & 26.93\%\\ 
 \hline

\end{tabular}}
\caption{\label{tab:3}Results of the \textcolor{black}{attack success rate}, accuracy before and after our attack, and attack decrease rate on each model and each dataset \textcolor{black}{(the accuracies reported above are only for 1000 randomly selected examples).}}
\end{table*}

\textcolor{black}{Furthermore, Table \ref{tab:3} summarizes the successfulness and effectiveness of our synonym BERT-based attack, measured by the Att.$_{\textsc{SR}}$ and Att.$_{\textsc{DR}}$ metrics, respectively.}
On the models level, our findings confirm that the DNN models (WordCNN and WordLSTM) are less susceptible to our attack than the BERT models.
For instance, the attack decrease rates of the WordCNN and WordLSTM models are both nearly 6\% on the MSDA dataset, whereas the Att.$_{\textsc{DR}}$ of the BERT model on the same dataset is approximately 27\%. 

On the other hand, on the datasets level, we observe that our synonym attack is more successful and less effective on the HARD dataset than the MSDA dataset, meaning our attack on the HARD dataset produced more adversarial examples than on the MSDA dataset, \textcolor{black}{but} at the same time, these newly produced adversarial examples exhibit less \textcolor{black}{impact} on the targeted models trained on the HARD dataset. \textcolor{black}{In contrast, our synonym attack generates} \textcolor{black}{fewer adversarial examples from the MSDA dataset, but those that succeed prove more potent.} \textcolor{black}{Notably, it is easier to successfully craft adversarial examples from the HARD dataset than the MSDA dataset since the MSDA dataset is \textcolor{black}{a Dialectal Arabic (DA)} dataset, and HARD is mostly \textcolor{black}{a Modern Standard Arabic (MSA)} dataset. Dialects in Arabic have fewer syntactic,  morphologic, and orthographic rules than official Modern Standard Arabic \citep{37}. } 

\section{Human Evaluation}
\label{sec:5}

\subsection{Setup of Human Evaluation}
We \textcolor{black} {invite} \emph{four} human evaluators (all native Arabic speakers) to evaluate the naturalness of the Arabic adversarial text examples generated by our synonym 
attack. We randomly \textcolor{black} {select} 150 adversarial text examples (50 examples for each model from the HARD dataset)\footnote{We only select generated adversarial examples from the HARD dataset because it is mostly written in Modern Standard Arabic, which can be easily evaluated in terms of Arabic grammar. MSA has syntactic,  morphologic, and orthographic rules, not like the Dialectical Arabic \citep{37}.} to be evaluated by our human evaluators in terms of two major criteria: grammatical similarity and semantic similarity. As an inner-level evaluation, we \textcolor{black}{ensure} that two of these native Arabic evaluators have college degrees in the Arabic language (linguists), while the other two do not (non-linguists), to study the inner difference in assessing the naturalness of our adversarial text examples between linguists and non-linguists\textcolor{black}{, and ask them to evaluate all the selected examples.}

 For grammatical similarity assessment, we first retrieve the corresponding original examples to the randomly selected adversarial examples and separately group them into two groups: original and adversarial. We then \textcolor{black}{task} all the human evaluators to rate both groups anonymously, meaning we do not tell them which group is which to guarantee that the original examples do not influence human evaluators’ judgment. Inspired by \citet{13}, we use a 5-point Likert scale, where \emph{one} represents \emph{strongly incorrect}, \emph{two} represents \emph{incorrect}, \emph{three} represents \emph{correct to some extent}, \emph{four} represents \emph{correct}, and \emph{five} represents \emph{strongly correct} \textcolor{black}{\citep{54}}. Following \citet{1}, we calculate the average score of the Likert scale \textcolor{black}{measurements} for each group (original and adversarial) and lastly divide the average score of the adversarial examples by the average score of the original examples of each human evaluator (linguists and non-linguists) to measure the precise grammatical similarity ratio between the original and adversarial examples. 

 For semantic similarity assessment, we accompany the randomly selected adversarial examples with their corresponding original examples and ask the human evaluators to rate whether the adversarial examples convey the same semantic meaning as the original examples. We use the same 5-point Likert scale, with different rating labels, ranging from \emph{one} representing \emph{strongly dissimilar} to \emph{five} representing \emph{strongly similar}. We then calculate the percentage of the average score of the Likert scale numbers (average score/number of rating labels) for each evaluator (linguists and non-linguists).

\begin{table*}[ht]
\centering
\tiny\renewcommand{\arraystretch}{0.90}
\resizebox{\textwidth}{!} {
\begin{tabular}{ccccc} 
 \hline
 \textbf{Evaluation Criteria} &  \textbf{Human Evaluator} &\textbf{WordCNN} & \textbf{WordLSTM} & \textbf{BERT}\\ 
  \hline
  & Linguists & 92.00\% & 94.00\% & 98.00\%\\ 
 \cline{2-5}
Grammatical Similarity  & Non-linguists & 99.00\% &  95.00\% & 98.00\%\\ 
\cline{2-5}
       &  {Overall Average} & 95.50\% &  94.50\% & 98.00\%\\ 
 \hline
   & Linguists & 89.00\% & 87.00\% & 91.00\%\\ 
 \cline{2-5}
    Semantic Similarity & Non-linguists & 87.00\% &  86.00\% & 86.00\%\\ 
 \cline{2-5}
     &  {Overall Average} & 88.00\% &  86.50\% & 88.50\%\\ 
 \hline
\end{tabular}}
\caption{\label{tab:4}Results of human evaluation of our generated adversarial text examples from each targeted model (WordCNN, WordLSTM, and BERT) on the HARD dataset; no examples used from the dialectical MSDA dataset.}
\vspace{-5pt}
\end{table*}

\subsection{Human Evaluation Results}
We observe in the grammatical similarity assessment that the non-linguist evaluators rated our adversarial examples and their original examples slightly higher than the linguist evaluators (except for the BERT model), whereas we have exactly the opposite results in the semantic similarity assessment (the linguists rated the examples higher than the non-linguists), as shown in Table  \ref{tab:4}. We assume that the gap in the background knowledge of the two human evaluators’ groups led to such results, where the non-linguists lack knowledge of Arabic syntax, making them rate the examples higher than the linguists in the grammatical similarity assessment. On the other hand, the deep understanding of the language and its semantics makes the linguists rate the examples higher than the non-linguists in the semantic similarity assessment.

Overall, as shown in Table \ref{tab:4}, the human evaluation results (across all 4 evaluators) demonstrate that the adversarial text examples generated by our synonym 
attack is acceptable to Arabic native speakers, meaning that our adversarial examples preserve a similar level of grammatical correctness and convey similar semantic meaning. For example, the overall average scores ranged from 94.50\% to 98\% for grammatical similarity, whereas the average scores ranged from 86.50\% to 88.50\% for semantic similarity.

\vspace{-5pt}

\section{Transferability}
\label{sec:6}
\vspace{-5pt}
\subsection{Setup of Transferability}
\vspace{-5pt}
The transferability of an adversarial attack refers to its ability to reduce the accuracy of the targeted models (victim models) to a certain extent when attacked by the newly generated adversarial examples from other models (source models), where all the victim and source models trained on the same dataset \citep{15, 14}. To closely examine the transferability, we examine 245 adversarial examples from each of the HARD and MSDA datasets along with their corresponding original examples. Initially, we predict the accuracy of both the original examples and their corresponding adversarial examples using the victim models. We then calculate the difference (delta $\Delta$) between the prediction accuracy of the original examples and the adversarial examples, ultimately determining the transferability score for each model. We have not seen any other researcher using this delta difference method for the transferability of adversarial text attacks, even in English.

\subsection{Transferability Results}
\vspace{-5pt}

Overall, first, we see that BERT (as the victim) has higher transferability scores than WordCNN or WordLSTM models, as shown in Table \ref{Table 5}. This indicates that BERT is more vulnerable to transferred attacks. This result is similar to what  \citet{1} saw in English. Second, we see that models trained in Dialectal Arabic (DA) are more vulnerable to transferred attacks than models trained in Modern Standard Arabic (MSA). 
\textcolor{black}{Again, fewer \textcolor{black}{generated} adversarial examples from the dialectical MSDA dataset satisfy the synonym rules, but those that succeed prove more potent \textcolor{black}{in the attack transferability between models.}}

\begin{table*}[ht]
\centering
\tiny\renewcommand{\arraystretch}{0.90}
\resizebox{\textwidth}{!} {
\begin{tabular}{ccccccccccc} 
\hline
   \multicolumn {11}{c}{\makecell{\textbf{Transferability Scores}}}\\
\hline
 \multirow{2}{*}{\makecell{\textbf{Datasets}}} & \multirow{2}{*}{\makecell{\textbf{Models}}} & 
 \multicolumn{3}{ c }{\makecell{\textbf{WordCNN}  \textbf{(source)}}} & \multicolumn{3}{ c }{\makecell{\textbf{WordLSTM}  \textbf{(source)}}} &\multicolumn{3}{ c }{\makecell{\textbf{BERT}  \textbf{(source)}}} \\ 
 \cline{3-11}
         &  & \makecell{$\textbf{X}$} & $\textbf{X}_{\textsc{\textbf{ADV}}}$ & $\Delta$ &  $\textbf{X}$  &  $\textbf{X}_{\textsc{\textbf{ADV}}}$ & $\Delta$ &  $\textbf{X}$  & $\textbf{X}_{\textsc{\textbf{ADV}}}$ & $\Delta$ \\  \cline{1-11}
         
        \multirow{3}{*}{\textbf{HARD}} & {\makecell{\textbf{WordCNN}  \textbf{(victim)}}} & --- & --- & --- & 52.65 &  47.34 & 5.31 & 65.71 & 34.28 & 31.43 \\  \cline{2-11}
       & {\makecell{\textbf{WordLSTM}  \textbf{(victim)}}} & 56.32 & 43.67 & 12.65 & --- &  --- & --- & 60.81 & 39.18 & 21.63 \\  \cline{2-11}

        &  {\makecell{\textbf{BERT}  \textbf{(victim)}}} & 75.51 & 24.48 & {\textbf{51.03}} & 74.28 &  25.71 & {\textbf{48.57}} & --- & --- & --- \\  \cline{1-11}
        \multirow{3}{*}{\textbf{MSDA}} & {\makecell{\textbf{WordCNN}  \textbf{(victim)}}} & --- & --- & --- & 87.34 &  12.65 & 74.69 & 86.53 & 13.46 & 73.07 \\  \cline{2-11}
       &{\makecell{\textbf{WordLSTM}  \textbf{(victim)}}} & 83.26 & 16.73 & 66.53 & --- &  --- & --- & 82.04 & 17.95 & 64.09 \\  \cline{2-11}

        & {\makecell{\textbf{BERT}  \textbf{(victim)}}} & 89.38 & 10.61 & {\textbf{78.77}} & 88.16 &  11.83 & {\textbf{ 76.33}} & --- & --- & --- \\ 
 \hline
\end{tabular}}
    \caption{Transferability of adversarial examples between victim and source models. Here, $X$ refers to original examples, and $X_{\textsc{ADV}}$ refers to adversarial examples. Let rows be $N$ and columns be $M$, then cells $N$$M$ are the accuracies of adversarial examples generated \textcolor{black}{from} the source model $M$ and evaluated on the victim model $N$.   A higher delta  $\Delta$ score indicates higher transferability between models.} 
    \label{Table 5}
\end{table*}

\section{Defense Mechanism}
\label{sec:7}

\subsection{Setup of Defense Mechanism}
We utilize adversarial training as a defense mechanism against our synonym 
attack, similar to \citet{1}, and follow the approach introduced by \citet{16}. This method is widely adopted in image classification to enhance models’ robustness. To assess whether the employment of adversarial training enhances the robustness of these models, we add the generated adversarial examples to the original datasets. Then, we retrain the models and evaluate the robustness of these newly adversarially trained models. We only study the effectiveness of the adversarial training on BERT models. 
We collect adversarial examples from both datasets that successfully fooled BERT models and incorporate them into the original datasets to train the models adversarially. We then refine-tune these BERT models using the \textcolor{black}{augmented} datasets with adversarial examples and subject these adversarially fine-tuned models to our attacks again.

\subsection{Defense Mechanism Results}
 We find that BERT models’ adversarial training accuracies increased on both datasets, compared to their accuracies after the attack with no defense mechanism in place, as seen in Table \ref{tab:6}. In other words, BERT models 
 \textcolor{black}{
 regain at least 2\% in accuracy} after applying adversarial training as a defense mechanism. These results agree with \citet{1}'s adversarial training results and provide further evidence that adversarial training is a good starting point for enhancing models’ robustness.

\begin{table}[h!]
\centering
\begin{tabular}{ccc} 
 \hline
 \textbf{Metric} & \textbf{HARD } & \textbf{MSDA} \\ 
  \hline
 Acc. Before Attack  & 88.59\% &  90.55\% \\ 
  \hline
 Acc. After Attack & 73.9\% & 63.62\%\\ 
 \hline
 Adversarial Training Acc. & 76.51\% & 65.69\% \\ 
 \hline
\end{tabular}
\caption{\label{tab:6}
Adversarial training results on BERT models.}
\end{table}

\vspace{-15pt}

 \section{Limitations}
 \label{sec:8}
\textcolor{black}{Due to the lack of availability of strong foundation models in Arabic and our lack of computational resources, one limitation of our attack is that it fundamentally depends on the pre-trained Arabic and multilingual models like 
the AraBERT model for retrieving synonym words, the  CAMeLBERT model for grammatical similarity checking, and the Sentence-Transformers model for semantic similarity checking, where their performance creates a bottleneck for the effectiveness of our text adversarial attack and the quality of our produced adversarial examples.} Another limitation of our work is that the successfulness of our attack in generating adversarial examples is higher than its \textcolor{black}{impact} on the targeted models because the evaluation misclassification rate of the targeted models is another bottleneck of our attack, meaning if a model’s original (evaluation) accuracy score is 86\% (e.g., BERT model on the MSDA dataset), then the misclassification rate of the model is 14\%, which undoubtedly affects the effectiveness of our attack negatively\textcolor{black}{, especially since our attack setting is black-box, where it is impossible to remove these misclassified examples without compromising the attack setting (if we remove these misclassified examples, the attack setting will be a white-box).}

\begin{table*}[ht]
\centering
\tiny\renewcommand{\arraystretch}{0.90}
\resizebox{\textwidth}{!} {
\begin{tabular}{cccc} 
 \hline
 \makecell{\textbf{Dataset}} & \textbf{Labels} & \textbf{Arabic Example} & \textbf{Translated Example}\\ 
  \hline
 \multirow{4}{*}{\makecell{\\ \\ \\ \\HARD}} & 
 Original Label: {Excellent} &   
\< ، فندق مريح جداً و كل شي كان جميل. > \textcolor{blue}{\< استثنائي >} & 
\textcolor{blue}{Exceptional} , very comfortable hotel and everything was beautiful.\\ 

\cline{2-4} & 
Attack Label: {Good} &   
\< ، فندق مريح جداً و كل شي كان جميل. > \textcolor{red}{\< ممتاز >} & 
\textcolor{red}{Excellent} , very comfortable hotel and everything was beautiful.\\

\cline{2-4} &
{Original Label}: Good &   
\makecell{\< ، استقبال جميل وفخامة ونظافة وسهولة> \textcolor{blue}{\< جيد >}  \\ \<  الوصول له وقريب من المطار والمراكز التسوق. >}& 

\makecell{\textcolor{blue}{Good} , beautiful reception, luxury, cleanliness, easy\\access, and close to the airport and shopping centers.}\\ 

  \cline{2-4}
     & {Attack Label}: Excellent &  
     \makecell{\< ، استقبال جميل وفخامة ونظافة وسهولة> \textcolor{red}{\< ممتاز >}  \\ \<  الوصول له وقريب من المطار والمراكز التسوق. >}& 

\makecell{\textcolor{red}{Excellent} , beautiful reception, luxury, cleanliness, easy\\access, and close to the airport and shopping centers.}\\ 
 \hline
 \multirow{4}{*}{\makecell{\\ \\ \\MSDA}}   & {Original Label}: Negative &   
\< نادين ويخليلك عيلتك. بعرف شو صعبة.> 
\textcolor{blue}{\< يرحم>}
\<الله>& \makecell{May God \textcolor{blue}{have mercy on} Nadine and keep \\your family. I know how difficult it is.}\\

  \cline{2-4}
     & {Attack Label}: Positive &   
\< نادين ويخليلك عيلتك. بعرف شو صعبة.> 
\textcolor{red}{\< يحمي>}
\<الله>&  \makecell{May God \textcolor{red}{protect} Nadine and keep \\your family. I know how difficult it is.}\\

  \cline{2-4}
     & {Original Label}: Positive &   
     \textcolor{blue}{\< يديمها.>}
     \<الأيام الحلوه كثير! يا صاحبي ربنا > & 
     Many good days! My friend, may God \textcolor{blue}{perpetuate them.}\\ 
  \cline{2-4}
     & {Attack Label}: Neutral &      
     \textcolor{red}{\< موجود.>}
     \<الأيام الحلوه كثير! يا صاحبي ربنا > & 
     Many good days! My friend, God \textcolor{red}{is always there.}\\  
 \hline

\end{tabular}}
\caption*{Appendix A:\hspace{4pt} Examples of the original Arabic and adversarial Arabic that are generated from the HARD and MSDA datasets and produced by the BERT models, along with their prediction labels and their English translations.} 
\label{Table 7}
\end{table*}

\section{Conclusion}
\label{sec:9}
In this work, we introduce the \textcolor{black}{\emph{first}} Arabic synonym 
\textcolor{black}{BERT-based} 
adversarial attack using an MLM task with pre-trained BERT models against the state-of-the-art text classification models. We utilize two large Arabic datasets, namely HARD and MSDA, in a black-box manner. We find that BERT and  \textcolor{black}{other} DNN models are generally susceptible to these Arabic adversarial examples, especially BERT models. We ask human evaluators to evaluate our produced adversarial examples using our attack in terms of grammatical similarity and semantic similarity and find that our attack generates examples that preserve semantic similarity and maintain Arabic grammar. We also study the transferability of these Arabic adversarial text examples by various source models on different victim models and observe that fine-tuned BERT models exhibit higher transferability when attacked by the \textcolor{black}{other} DNN models’ generated adversarial examples. Lastly, we investigate the effectiveness of the adversarial training defense mechanism on BERT models and find that the BERT models successfully regain at least 2\% of their accuracies after applying the adversarial training as a defense mechanism.

\section*{Reproducibility}
\textcolor{black}{We share our code scripts and trained models on GitHub at \url{https://www.github.com/NorahAlshahrani/bert_synonym_attack}.}

\section*{Acknowledgments}
\textcolor{black}{We would like to thank Clarkson University and the Office of Information Technology (OIT) for providing computational resources and support that contributed to these research results.} \textcolor{black}{We also thank the anonymous evaluators who participated in the human evaluation study.}

\bibliography{anthology,custom}
\bibliographystyle{acl_natbib}

\end{document}